\documentclass{Interspeech2024}

\newenvironment{tight}{%
\setlength{\abovedisplayskip}{3pt}
\setlength{\belowdisplayskip}{3pt}
}

\usepackage{CJKutf8}
\usepackage{CJK}
\usepackage{tabularx}

\interspeechcameraready
\title{A Multitask Training Approach to Enhance Whisper\\ with Open-Vocabulary Keyword Spotting}

\name{Yuang Li, Min Zhang, Chang Su, Yinglu Li, Xiaosong Qiao, Mengxin Ren, \\ Miaomiao Ma, Daimeng Wei, Shimin Tao, Hao Yang}

\address{
  Huawei Translation Services Center, China
\email{\{liyuang3, zhangmin186, suchang8, liyinglu, qiaoxiaosong, renmengxin2, mamiaomiao, weidaimeng, taoshimin, yanghao30\}@huawei.com}}

\keywords{speech recognition, keyword spotting, prompt}

\begin{document}

\maketitle

\begin{abstract}
The recognition of rare named entities, such as personal names and terminologies, is challenging for automatic speech recognition (ASR) systems, especially when they are not frequently observed in the training data. In this paper, we introduce keyword spotting enhanced Whisper (KWS-Whisper), a novel ASR system that leverages the Whisper model and performs open-vocabulary keyword spotting (OV-KWS) on the hidden states of the Whisper encoder to recognize user-defined named entities. These entities serve as prompts for the Whisper decoder. To optimize the model, we propose a multitask training approach that learns OV-KWS and contextual-ASR tasks. We evaluate our approach on Chinese Aishell hot word subsets and two internal code-switching test sets and show that it significantly improves the entity recall compared to the original Whisper model. Moreover, we demonstrate that the OV-KWS can be a plug-and-play module to enhance the ASR error correction methods and frozen Whisper models.

\end{abstract}

\section{Introduction}

Recent years have witnessed the rise of end-to-end (E2E) automatic speech recognition (ASR) models~\cite{chorowski2015attention,graves2012sequence,graves2006connectionist} due to their simplicity and unified architecture. However, these models struggle to recognize proper nouns that infrequently occur in the training data. A common approach is to construct a list of hot words and apply it to shallow fusion~\cite{Adi_2021, Zhao_2019} or deep fusion~\cite{Chang_2021,Chen_2019,han_2022,Tsen_2022,tara_2023}. Shallow fusion enhances the hot words' score during beam search decoding, but it requires tuning the optimal fusion weight. The deep fusion method trains a contextual encoder jointly with the ASR model from scratch. These methods map the entity words and the speech signal into the same feature space, and the decoder leverages both contextual and acoustic information to generate the transcription. To adapt existing ASR models, some methods use adaptors~\cite{Sa_2022,Tong_2023} to modify the intermediate features of the ASR model or pointer networks~\cite{Sun_2023} to revise the output distributions. Moreover, the shallow and deep fusion methods can be combined to enhance the performance~\cite{Le_2021,xU_2023}.

Whisper~\cite{radford2022robust} stands as a cutting-edge ASR model. It leverages the Transformer architecture~\cite{NIPS2017_3f5ee243} and was trained on an extensive dataset comprising 680k hours of speech data. Unlike its predecessors, which necessitate a separate biasing module, Whisper dynamically adjusts its output by incorporating a straightforward prompt during decoding. In this paper, we introduce an innovative technique called KWS-Whisper (Keyword Spotting Enhanced Whisper). This method integrates an open-vocabulary keyword spotting (OV-KWS) module between the encoder and decoder of Whisper. The OV-KWS module draws inspiration from vision-based keyword spotting methods~\cite{momeni2020seeing,audio_text} and constructs a cosine similarity matrix using features from the Whisper encoder. To enhance Whisper’s comprehension of prompts, we employ a multitask training approach that combines OV-KWS and contextual-ASR tasks. Our system enhances entity recall with absolute improvements of up to 80\% on Aishell hot word subsets~\cite{aishell_subset} and up to 10\% on internal code-switching datasets. However, we noticed that the KWS-Whisper fine-tuned on a small-scale dataset has higher mixed error rates (MERs) than the Whisper model on real-world internal datasets due to catastrophic forgetting. Therefore, we propose to use OV-KWS as an independent module for frozen Whisper models. We conducted experiments for various scenarios: 1) ASR error correction based on Pinyin or the large language model (LLM); 2) directly prompting the Whisper decoder with spoken-form prompts that resemble the historical transcription. The results show that our system can achieve significant MER reductions of 2.4\%, 1.4\%, and 1.8\% on internal datasets for Whisper-small, medium, and large models respectively. Compared to our previous work~\cite{li-etal-2024-cb-whisper}, the lightweight OV-KWS module and the multitask training approach markedly enhance the system’s adaptability, efficiency, and accuracy.

\begin{figure}[t]
  \centering
  \includegraphics[width=0.8\linewidth]{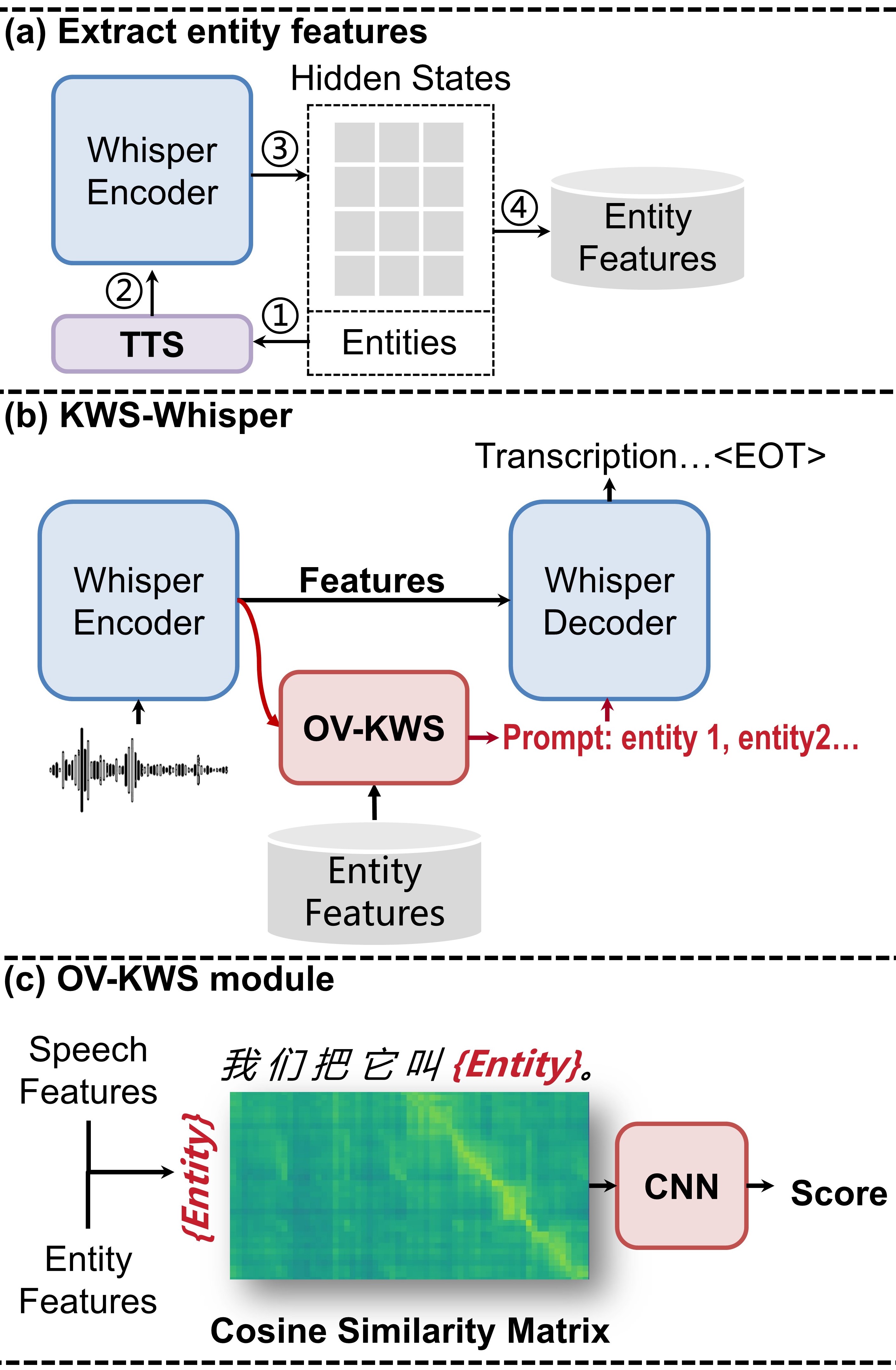}
  \caption{(a) Entity Features are extracted using TTS followed by Whisper encoder. (b) Flowchart of the KWS-Whisper model. (c) Illustration of the OV-KWS module.}
  \label{fig:sys}
\end{figure}

\section{Methodology}

\subsection{System overview}

A database is first constructed by extracting acoustic representations for pre-defined entity words. As shown in Figure~\ref{fig:sys} (a), a text-to-speech (TTS) model followed by the Whisper encoder is used to extract hidden states, which are later matched with the input utterance features. As illustrated in Figure~\ref{fig:sys} (b), the proposed KWS-Whisper identifies entity words before decoding by leveraging encoder hidden states. Specifically, we aggregate hidden representations from all Whisper encoder layers, weighted by learnable coefficients. A cosine similarity matrix is then computed between the input utterance features and stored entity word features. A binary Convolutional Neural Network (CNN) detects the presence of an entity word by recognizing the diagonal pattern in the similarity matrix (Figure~\ref{fig:sys} (c)). Finally, predicted entity words guide the Whisper decoder for improved recognition accuracy.

\subsection{Open-vocabulary keyword spotting}

OV-KWS is framed as a binary classification task. The model predicts the presence of an entity word within an utterance by analyzing the cosine similarity matrix between the hidden states of entity words and the input utterance. Notably, if an entity word exists in the utterance, a diagonal pattern emerges (Figure~\ref{fig:sys} (c)). Given that this is a local pattern, we opt for a CNN as our classification model. To extract features, we follow a common approach of using the weighted sum of multi-layer features from the Whisper encoder~\cite{Chen_2022}. The OV-KWS module exhibits linear computational complexity to the number of keywords. For higher efficiency, we employ an ultra-lightweight CNN comprising only four layers and 0.2 million parameters. Additionally, we leverage GPU parallelization to batch and execute multiple keywords simultaneously. An easy way to construct the dataset for the CNN classifier is to select positive words according to the transcription and choose negative words randomly. However, this approach is too simplistic, as the negative samples may have significant differences from the positive samples, resulting in overfitting. As such, we adopt hard negative samples~\cite{navon2023openvocabulary}, which are more likely to have overlaps with positive samples, increasing the difficulty of the classification task. 

\subsection{Multitask fine-tuning}

The CNN classifier's capacity is limited by its small size. Therefore, it is necessary to fine-tune the Whisper model for the OV-KWS task while preserving its ASR ability. To this end, we propose to fine-tune all parameters of the Whisper model on both the OV-KWS task and the contextual-ASR task jointly with the loss function shown in Equation~\ref{eq:1}. During fine-tuning, we incorporate words from the ground-truth transcription into the prompt following the naive prompt format in Table~\ref{table:prompt}. This enables the Whisper decoder to better understand the prompt format and leverage the word in the prompt.

\begin{tight}
\begin{align}
\label{eq:1}
Loss = \mathcal{L}_{ASR} + \alpha \times \mathcal{L}_{KWS}
\end{align}
\end{tight}

\noindent where the loss is the weighted sum of ASR loss $\mathcal{L}_{ASR}$ and KWS classification loss $\mathcal{L}_{KWS}$. The weight $\alpha$ was set to three during training.

\begin{CJK}{UTF8}{gbsn}
\begin{table}[t]
\small
\caption{Prompts for the Whisper decoder.}
    \centering
    \begin{tabular}{p{1.7cm} | p{4cm}} % Adjust the width of the second column as per your requirement
    \toprule
    naive prompt & 实体1, 实体2, 实体3\\
    
    (translation) & entity 1, entity 2, entity 3\\
    \midrule
    \textbf{spoken-form prompt} & 今天演讲的主题是这个呃，实体1、实体2、实体3。好，那我就继续讲。\\
    (translation) &  The topic of today's speech is, ah, entity 1, entity 2, entity 3. Okay, then I'll continue.\\
    \bottomrule
    \end{tabular}
    \label{table:prompt}
\end{table}
\end{CJK}

\subsection{Prompting Whisper decoder}

During training, the Whisper decoder incorporates contextual information by using the transcription of the previous utterance as a prompt. In inference, any relevant text-only prompt can serve as the historical context for decoding. For multitask training, we adopt a straightforward prompt format (i.e., concatenating all entity words) which performs remarkably well on in-domain test sets (Aishell). However, this format can lead to hallucinations in more complex spontaneous conversations involving code-switching. Additionally, the fine-tuned model's ASR performance may slightly degrade due to its small-scale training dataset. Therefore, we explore using OV-KWS as a plug-in for correcting ASR errors and prompting frozen Whisper models.

\subsection{OV-KWS plug-in for frozen Whisper models}

We can use the OV-KWS plug-in for:

\begin{itemize}
    \item \textbf{ASR error correction}: 1) Pinyin-based method: We segment each transcription into words and replace the Chinese word if it has the same Pinyin (pronunciation) as the entity word. For English words, we compute the edit distance and replace the word if the distance is below a pre-defined threshold. 2) LLM-based method: We use ChatGLM3-6B~\cite{zeng2023glm130b} and the prompt in Table~\ref{table:prompt_llm}. In our experiments, we provide two examples for the LLM.
    \item \textbf{Prompting Whisper decoder}: Since the Whisper models are not fine-tuned, we use spoken-form prompt (Table~\ref{table:prompt}) to prevent hallucinations which incorporates disfluencies and indicate that the subsequent speech was a spontaneous talk. Compared to the naive prompt, the spoken-form prompt has a more similar form to the historical transcriptions used during the training of the Whisper model.
\end{itemize}

\begin{CJK}{UTF8}{gbsn}
\begin{table}[h]
\small
\caption{The prompt for LLM-based ASR error correction. The translation of the Chinese prompt is given in parentheses with italics.}
    \centering
    \begin{tabular}{p{7cm}} % Adjust the width of the second column as per your requirement
    \toprule
    \textbf{User}: \\请根据可能出现的关键词修改ASR文本。\\
    (\textit{Please modify the ASR transcription according to possible keywords.})\\
    关键词 (\textit{keywords})：\{word1\}，\{word2\}。\\
    识别文本 (\textit{recognized text})：\{ASR transcription\}\\
    \textbf{Response}: \\修改文本 (\textit{modified text}): \\
    
    \bottomrule
    \end{tabular}
    \label{table:prompt_llm}
\end{table}
\end{CJK}

\section{Experimental Setups}

\subsection{Training configurations}

We selected the Whisper-small model for the multitask training. In subsequent experiments, we also compared the Whisper-small, medium, and large-v2 versions without fine-tuning. For the CNN, we employed a lightweight four-layer architecture. Each layer had 128, 128, 256, and 256 channels respectively, and a kernel size of three. In the first training phase, we froze the Whisper encoder and trained the CNN classifier for OV-KWS and the Whisper decoder for ASR, for 10 epochs, with a batch size of 128 and a learning rate of $5×10^{-5}$. The total number of trainable parameters was 153.8 million (CNN 0.2 million + decoder 153.6 million). In the second training phase, we unfroze the Whisper encoder and trained the entire model for another 50 epochs, with a learning rate of $3×10^{-5}$. The number of trainable parameters increased to 241.9 million. Note that for the ASR task, we applied a random concatenation strategy~\cite{yu2023connecting} to increase the length of the utterance. We used a beam size of five for decoding.

\subsection{Datasets}

To construct the training set for OV-KWS and ASR, we utilized three datasets: Aishell-1~\cite{aishell}, LibriSpeech~\cite{Panayotov_2015}, and TALCS~\cite{li2022talcs}. Aishell-1 is a Mandarin ASR dataset with 150 hours of speech recordings. LibriSpeech is an English ASR dataset, from which we selected the train-clean subset containing 100 hours of speech recordings. TALCS is a Mandarin-English code-switching dataset with 578 hours of speech data from English teaching scenes. For each utterance in these datasets, we randomly sampled positive words and generated speech using edge-tts~\footnote{\url{https://github.com/rany2/edge-tts}}. We compared the performance of the model trained with only Chinese data (zh) and the model trained with the mixture of three datasets (zh + en).

\begin{table}[h]
\small
\caption{Statistics of ASR test sets.}
\label{tab:dataset}
\centering
\begin{tabular}{c | c c c} 
\toprule
Dataset & Utterances & Duration (min) & Entities\\
\midrule
Aishell-dev & 1334 & 113 & 371 \\
Aishell-test & 808 & 76 & 226 \\
Internal-1 & 99 & 41 & 150 \\
Internal-2 & 112 & 47 & 346 \\
\bottomrule
\end{tabular}
\end{table}

We evaluated the performance of OV-KWS and ASR on four different test sets, which consisted of two internal datasets and two open-source datasets. The internal datasets contain technical talks and manually labeled entity words that are mainly in Chinese but also include some English terms. The open-source datasets are the Aishell hot word subsets~\cite{aishell_subset} in Chinese. The statistics of these datasets are shown in Table~\ref{tab:dataset}.

\subsection{Evaluation metrics}

We evaluated the performance of OV-KWS using the F1 score, which is the harmonic mean of the precision and recall. For ASR, we adopted two metrics: mixed error rate (MER) and entity recall. MER is a modified version of character error rate (CER) that can handle code-switching situations between English and Chinese. In this metric, we treat each Chinese character and each English word as a single unit. Therefore, MER is equivalent to CER for Chinese and word error rate (WER) for English.

\section{Experimental Results}
\subsection{Results for open-vocabulary keyword spotting}

\begin{figure}[h]
  \centering
  \includegraphics[width=\linewidth]{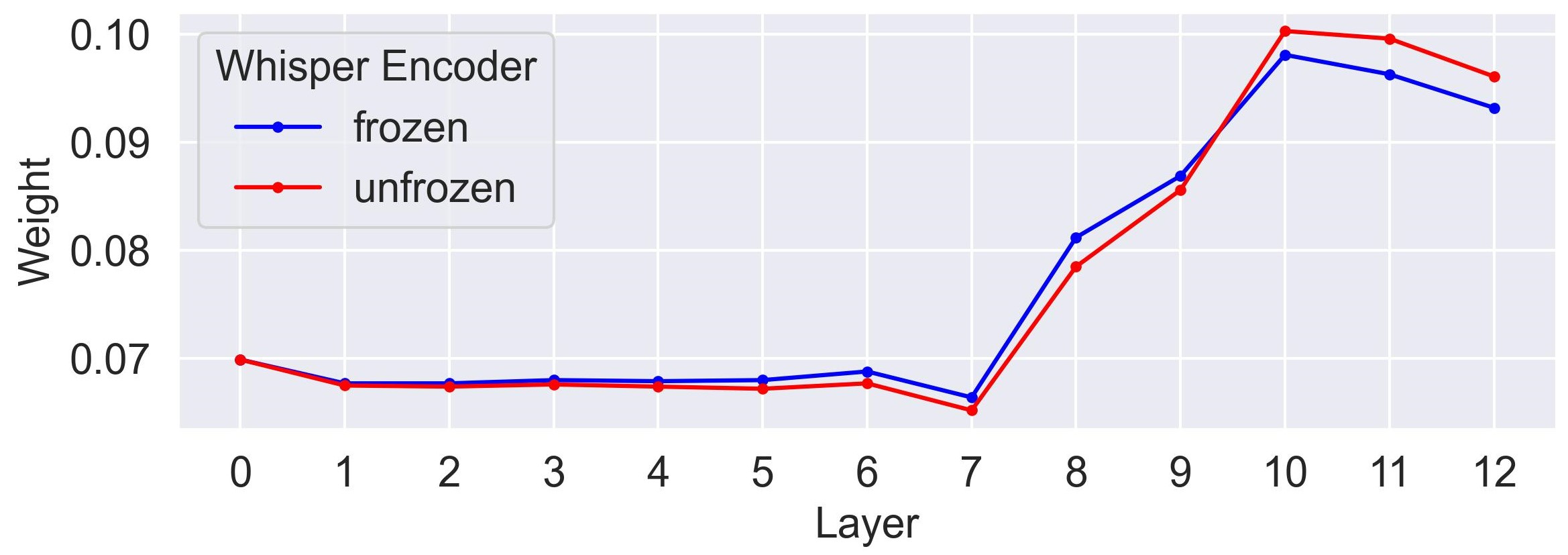}
  \caption{The weights for Whisper encoder layers.}
  \label{fig:weight}
\end{figure}

We evaluated the performance of the CNN classifier for OV-KWS using the F1 score. Table~\ref{tab:kws} shows the results of different training settings. We first trained the model with a frozen Whisper encoder on the Aishell dataset only and obtained F1 scores of around 0.6 on the Aishell test sets and 0.68 on the internal datasets. Then, we evaluated the model trained with a mixture of three datasets and observed consistent improvement in the F1 scores. Finally, we unfroze the Whisper encoder, which increased the number of parameters of the OV-KWS task from 0.2 million to 88.4 million and achieved significant improvement in the F1 scores. The model trained with the Aishell dataset performed slightly better on the Aishell-dev and Aishell-test sets, with F1 scores of 0.85 and 0.89, respectively. On the internal datasets, the model trained with the mixed dataset achieved higher F1 scores, reaching around 0.9. The comparison between single-task training and multitask training is presented in Table~\ref{tab:ablation}. The results indicate that simultaneous optimization of ASR and OV-KWS does not adversely affect the performance of the OV-KWS task instead, it leads to a higher F1 score on the Internal-1 dataset. Figure~\ref{fig:weight} illustrates the weight assigned to each layer of the Whisper encoder. The hidden states from the 8th to the 12th layers have a greater impact on the OV-KWS task. Moreover, the weight for the deeper layers increased slightly after fine-tuning, suggesting that the Whisper encoder is more adapted to the OV-KWS task.

\begin{table}[h]
\small
\caption{The performance of the OV-KWS module measured by \textbf{F1 score}. We compared the models trained with Chinese-only data (zh) and Chinese-English-code-switching data (zh + en). We also compared the results with the Whisper model frozen or unfrozen.}
\label{tab:kws}
\centering
\begin{tabular}{c | c c | c c } 
\toprule
 Training data & \multicolumn{2}{c|}{zh} & \multicolumn{2}{c}{zh + en}\\
 Unfreeze Whisper & × & \checkmark & × & \checkmark \\
\midrule
Aishell-dev & 0.62 & \textbf{0.85} & 0.64 & 0.83 \\
Aishell-test & 0.60 & \textbf{0.89} & 0.61 & 0.84 \\
Internal-1 & 0.68 & 0.75 & 0.71 & \textbf{0.90} \\
Internal-2 & 0.68 & 0.82 & 0.72 & \textbf{0.88} \\
\bottomrule
\end{tabular}
\end{table}

\begin{table}[h]
\small
\caption{The comparison between single-task training and multitask training of the OV-KWS model measured by \textbf{F1 score}.}
\label{tab:ablation}
\centering
\begin{tabular}{c | c c} 
\toprule
 & OV-KWS & OV-KWS + ASR\\
\midrule
Aishell-dev & \textbf{0.86} & 0.83\\
Aishell-test & \textbf{0.86} & 0.84\\
Internal-1 & 0.87 & \textbf{0.90}\\
Internal-2 & \textbf{0.88} & \textbf{0.88}\\
\bottomrule
\end{tabular}
\end{table}

\begin{table*}[t]
% \small
\caption{The performance of KWS-Whisper measured by \textbf{MER (\%) / Entity Recall (\%)}. The Whisper-small model was unfrozen and trained jointly with the OV-KWS module. "zh" means the model was fine-tuned on Chinese-only data. "zh + en" represents the model was trained on Chinese-English-code-switching data. We prompted the Whisper decoder with ground-truth entities and predicted entities respectively. }
\label{tab:asr1}
\centering
\begin{tabular}{c | c c | c c c | c c c} 
\toprule
Model & \multicolumn{2}{c|}{Whisper-small (baseline)} & \multicolumn{3}{c|}{KWS-Whisper-small (zh)} & \multicolumn{3}{c}{KWS-Whisper-small (zh + en)}\\
 Prompt & × & ground-truth & × & ground-truth & prediction & × & ground-truth & prediction\\
\midrule
Aishell-dev & 16.7 / 6.3 & 13.3 / 81.6 & 10.4 / 20.9 & 7.2 / 87.1 & \textbf{7.6 / 84.2} & 12.1 / 19.6 & 8.5 / 91.5 & \textbf{9.2 / 88.4} \\
Aishell-test & 20.1 / 10.1 & 15.9 / 82.1 & 11.6 / 14.0 & 8.3 / 84.4 & \textbf{8.6 / 82.4} & 13.5 / 9.8 & 9.7 / 91.6 & \textbf{10.2 / 87.7} \\ 
Internal-1 & 5.9 / 78.5 & 5.3 / 95.3 & 14.7 / 49.0 & 13.5 / 76.7 & 13.9 / 70.2 & 8.9 / 61.7 & 8.7 / 86.8 & 7.7 / \textbf{86.3} \\
Internal-2 & 9.7 / 60.8 & 8.9 / 93.7 & 20.0 / 32.5 & 17.4 / 69.5 & 17.2 / 64.8 & 16.6 / 35.9 & 13.0 / 76.4 & 13.2 / \textbf{71.4}\\
\bottomrule
\end{tabular}
\end{table*}

\subsection{Results for KWS-Whisper}

The multitask training of our KWS-Whisper model yielded the ASR performance shown in Table~\ref{tab:asr1}. Compared to the original Whisper-small model, the fine-tuned KWS-Whisper model achieved a significantly lower MER and higher entity recall on the Aishell-dev and Aishell-test sets. The model that was trained on the mixed dataset exhibited higher entity recall across all datasets, indicating better prompting function and generalizability. Remarkably, the KWS-Whisper model with predicted entities surpassed the Whisper-small model with ground-truth entities in entity recall, reaching 88.4\% and 87.7\% on the Aishell-dev and test sets respectively. Furthermore, the KWS-Whisper model outperformed the state-of-the-art SeACo-Paraformer model~\cite{aishell_subset}, which has entity recalls of 86\% and 87\% on the Aishell-dev and Aishell-test sets. 

However, we found that though the entity recalls could be enhanced on internal test sets, the MERs increased slightly. This was mainly because we fine-tuned KWS-Whisper on small-scale datasets, and the domain of the training data (audiobook and teaching scenes) differed from that of the test sets (technical talks). As a result, Whisper’s generalizability was reduced. Hence, we suggest using OV-KWS as a plug-in for ASR error correction or applying frozen Whisper models. Table~\ref{tab:asr2} shows that error correction with Chinese Pinyin, using the recognized entity from the OV-KWS model, consistently improved the MER and entity recalls for all Whisper model sizes. The LLM-based correction achieved higher entity recalls than the naive Pinyin-based method, as it incorporated contextual information. However, the MER increased due to the LLM's hallucinations. We also employed the prompting approach in KWS-Whisper, without fine-tuning various Whisper models. This method outperformed error correction–based methods significantly, as it considered acoustic features. Nevertheless, we discovered that the naive prompt slightly raised the MER compared to the original model. The main reason was that the Whisper model tends to omit filler words and disfluencies if the prompt is in well-formatted text. This problem could be alleviated by using the spoken-form prompt that mimics a presentation transcription. Compared to the naive prompt and no prompt, spoken-form prompts consistently improved the MER. For example, using the OV-KWS model and the spoken-form prompt, the MER decreased from 4.4\% to 2.9\% and from 5.6\% to 3.6\% on the Internal-1 and Internal-2 subsets respectively for the Whisper-large-v2 model. In addition, the best system attained entity recalls of 97.7\% and 96.9\% for the two datasets, considerably superior to the original model's performance of 87.3\% and 79.9\%, respectively.

\begin{table}[t]
% \small
\caption{The performance of various systems with OV-KWS as an independent module measured by \textbf{MER (\%) / Entity Recall (\%)}. We investigated the effects of two ASR error correction techniques: Pinyin-based and LLM-based, and two types of prompts for the Whisper decoder: naive and spoken-form.}
\label{tab:asr2}
\centering
\begin{tabular}{c | c | c c} 
\toprule
Model & Method & Internal-1 & Internal-2\\
\midrule
Whisper- & × & 5.9 / 78.5 & 9.7 / 60.8 \\
small & Correction-Pinyin & 5.7 / 82.9 & 9.1 / 68.2 \\
 & Correction-LLM & 6.9 / 85.2 & 12.7 / 74.3 \\
 & Prompt-Naive  & 5.4 / 94.0 & 9.0 / 90.2 \\
 & Prompt-Spoken & \textbf{4.0 / 95.9} & \textbf{6.8 / 91.1} \\
\midrule
Whisper & × & 4.8 / 84.2 & 6.7 / 70.4 \\
-medium & Correction-Pinyin & 4.8 / 85.2 & 6.3 / 78.3 \\
 & Correction-LLM & 6.1 / 88.9 & 9.3 / 80.9 \\
 & Prompt-Naive & 6.8 / 94.3 & 14.6 / 90.3 \\
& Prompt-Spoken & \textbf{4.0 / 95.6} & \textbf{4.8 / 94.4} \\
\midrule
Whisper & × & 4.4 / 87.3 & 5.6 / 79.9 \\
-large-v2 & Correction-Pinyin & 4.3 / 88.9 & 5.5 / 83.9 \\
 & Correction-LLM & 6.0 / 89.9 & 7.7 / 85.1 \\
 & Prompt-Naive & 4.7 / 97.4 & 6.6 / 95.1 \\
& Prompt-Spoken & \textbf{2.9 / 97.7} & \textbf{3.6 / 96.9} \\ 
\bottomrule
\end{tabular}
\end{table}

\section{Conclusion}

This paper presents KWS-Whisper, an innovative ASR framework that leverages the prior knowledge of entity words to enhance their recalls. Our model consists of an OV-KWS module that exploits the hidden states of the Whisper encoder. The entities detected by this module serve as prompts for the Whisper decoder. We devise a multitask training strategy that simultaneously optimizes OV-KWS and contextual-ASR tasks. We also explore more convenient methods to apply the OV-KWS model to correct ASR errors and to prompt frozen Whisper models. Evaluated on four test sets, our method is shown to achieve substantial improvements in both MER and entity recall. In future work, we plan to enhance the efficiency of the OV-KWS module and enable it to handle a large keyword database and conduct large-scale experiments by distilling speech data from the original Whisper model, thereby boosting the performance of multitask fine-tuning in the general domain.

\clearpage

\bibliographystyle{IEEEtran}
\bibliography{mybib}

\end{document}